\pdfoutput=1

\documentclass[11pt]{article}
\usepackage[preprint]{acl}
\usepackage{times}
\usepackage[T1]{fontenc}
\usepackage[utf8]{inputenc}
\usepackage{microtype}
\usepackage{graphicx}
\usepackage{booktabs}
\usepackage{array}
\usepackage{float}
\usepackage{listings}
\lstdefinestyle{prompt}{basicstyle=\footnotesize\ttfamily, columns=fullflexible, keepspaces=true, upquote=true,
  breaklines=true, breakatwhitespace=true, breakindent=1.5em, xleftmargin=1em, aboveskip=3pt, belowskip=3pt}

\newcommand{\dsname}{\textsc{CompOrca}}
\newcommand{\judge}{LongCat-2.0}
\newcommand{\judgeb}{DeepSeek-V4-Flash}

\title{\dsname: Corpus-Scale Compliance Labelling of Instruction-Tuning Data\thanks{\;Accepted to the PlurVA-LLM Workshop at AACL-IJCNLP 2026.}}

\author{
  Philipp E.\ Glass\thanks{\;Correspondence: \texttt{phil.glass@cemiu.net}} \quad
  Alina Miron \\
  Department of Computer Science, Brunel University of London \\
  \texttt{\{phil.glass, alina.miron\}@brunel.ac.uk}
}

\begin{document}
\maketitle

\begin{abstract}
Studying how fine-tuning shapes refusal and noncompliance behaviour requires knowing which training examples refuse or otherwise fail to fulfil the request. Existing annotations cover evaluation sets, which are far smaller than training corpora. We present \dsname, compliance labels for all 4{,}233{,}923 examples of the OpenOrca corpus. Every example was classified as compliant or noncompliant by five passes of an open-weight LLM judge (LongCat-2.0, 1.6T parameters). The corpus is released as unanimous compliance (94.75\%), unanimous noncompliance (1.28\%), and nonunanimous rows (3.97\%), with the raw vote counts. A single pass flags 2.7--3.2\% of the corpus as noncompliant, while only 1.28\% is flagged by all five, so the most ambiguous rows can be filtered out. Against 450 human-annotated examples (150 annotated twice; human--human $\kappa=0.93$), the unanimous compliance and noncompliance labels are 97.3\% and 86.7\% precise. The noncompliance label is a high-precision subset of the corpus's noncompliance. Published refusal-detection methods recall between 0.4\% and 94.1\% of human-labelled noncompliance. We release the full corpus with its per-row labels and vote counts at \url{https://huggingface.co/datasets/cemiu/CompOrca}.
\end{abstract}

\section{Introduction}
\label{sec:intro}

In LLMs, refusal is mediated by a direction in activation space \citep{arditi2024refusal}, targeted by automated attacks \citep{zou2023universal,mazeika2024harmbench}, prone to over-refusal \citep{rottger2024xstest,cui2025orbench}, and trainable from noncompliance data \citep{brahman2024coconot,bianchi2024safetytuned}. Much of this work concerns training data, as fine-tuning can erode the behaviour or bring it back \citep{qi2024finetuning}. Studying it requires knowing which examples in a corpus do not answer the request, for example to build mixtures with controlled contamination, to ablate such rows or to track what a model was trained on. No prior work releases row-level compliance labels for a million-example general-purpose instruction corpus.

Refusal labels are also often too narrow for these uses. Refusal describes an assistant declining to act, while instruction corpora mostly contain \emph{noncompliance} examples. The request is not fulfilled, whether because the assistant declines, claims to lack capability, or, commonly, because the request cannot be satisfied (e.g.\ the passage does not contain the answer, the question is malformed). We label this broader class, following \citet{brahman2024coconot}, which significantly influences which rows count as noncompliant.

\begin{figure}[t]
\centering
\includegraphics[width=\columnwidth]{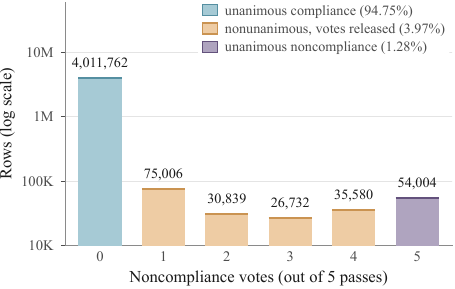}
\caption{\dsname{} vote distribution over the 4.23M OpenOrca rows (log scale). Each pass flags 2.7--3.2\% of rows as noncompliant; 1.28\% is flagged by all five. Most nonunanimous rows differ only by a single vote.}
\label{fig:votes}
\end{figure}

The default corpus-scale tool is substring matching against common refusal phrases (``Sorry, I can't''), built for scoring jailbreak attacks \citep{zou2023universal}. On an instruction corpus these lists fail in both directions, and \citeauthor{zou2023universal}'s list reaches only 12.6\% precision and 6.3\% recall against our labels (Section~\ref{sec:baselines}). LLM-as-a-judge \citep{zheng2023judging} is more accurate, but unstable. Across five independent passes over the full corpus, 52--60\% of examples flagged as noncompliant by a single pass are not flagged by all five (Section~\ref{sec:reliability}). We release \dsname, a compliance labelling of the full OpenOrca corpus \citep{lian2023openorca}. Our contributions are:

\begin{itemize}
\item \textbf{Dataset.} The corpus in full with per-example labels and raw vote counts: 4.01M unanimous-compliance rows, a stricter 3.77M-row compliance-clean subset, 54K unanimous-noncompliance rows, and 168K nonunanimous rows.

\item \textbf{Validation and comparison.} Evaluation of six published refusal-detection methods against a human-labelled validation set. Their recall on human-labelled noncompliance ranges from 0.4\% to 94.1\%, against 97.0\% for \dsname{} on the rows it labels.

\end{itemize}

\section{Related Work}
\label{sec:related}

\paragraph{Refusal and noncompliance.}
Existing resources are prompt evaluation sets or moderation models: XSTest \citep{rottger2024xstest}, OR-Bench \citep{cui2025orbench}, SORRY-Bench \citep{xie2025sorrybench}, FalseReject \citep{zhang2025falsereject}, and CoCoNot \citep{brahman2024coconot}, whose noncompliance framing we adopt. Among classifiers, WildGuard \citep{han2024wildguard} includes response-refusal detection and Do-Not-Answer \citep{wang2024donotanswer} trained refusal classifiers. Closest to this work, \citet{recum2024cannot} build a taxonomy of refusals from 8.6K annotated instances in IFT/RLHF corpora. These resources contain fewer than $10^5$ examples. None labels a whole, existing training corpus.

\paragraph{Substring detection.}
Most work detects refusal with \citeauthor{zou2023universal}'s (\citeyear{zou2023universal}) thirteen refusal prefixes, often reused unchanged \citep{chao2024jailbreakbench}. The list was built for scoring jailbreaks, but was shown to be unreliable even there \citep{mazeika2024harmbench,souly2024strongreject}. We measure how poorly it transfers to an instruction corpus.

\paragraph{LLM-as-a-judge.}
On many annotation tasks, LLM judges agree with humans nearly as well as humans agree with each other \citep{zheng2023judging,gilardi2023chatgpt}, but they are biased \citep{wang2024fair} and vary run to run \citep{yagubyan2026coinflip}. For reasoning, the usual remedy is to sample several answers and aggregate them \citep{wang2023selfconsistency}, and we do the same for annotation.

\section{Dataset Construction}
\label{sec:dataset}
\begin{table}[t]
\centering\small
\begin{tabular}{@{}lrr@{}}
\toprule
\textbf{Bucket} & \textbf{Rows} & \textbf{Share (\%)} \\
\midrule
Unanimous compliance (0/5) & 4{,}011{,}762 & 94.75 \\
Nonunanimous (1--4/5) & 168{,}157 & 3.97 \\
Unanimous noncompliance (5/5) & 54{,}004 & 1.28 \\
\midrule
Compliance-clean subset & 3{,}769{,}249 & 89.02 \\
\midrule
Total & 4{,}233{,}923 & 100.00 \\
\bottomrule
\end{tabular}
\caption{Release structure of \dsname{}, by the number of the five judge passes that label a row noncompliant.}
\label{tab:buckets}
\end{table}

\paragraph{Source corpus.}
We label the \texttt{train} split of \texttt{Open-Orca/OpenOrca}: 4{,}233{,}923 rows, roughly 1M GPT-4 and 3.2M GPT-3.5 augmentations of FLAN tasks.

\paragraph{Label definition.}
A row is noncompliant (the positive class) if the requested content is not delivered. The judge prompt (Appendix~\ref{app:prompt}) asks for \textsc{refusal} when a response ``declines to fulfill the request, in whole or in substantial part'', including refusals on capability grounds and ``deflections that withhold the requested content'', and for \textsc{compliance} when it ``attempts to fulfill the request, even partially, even with caveats or disclaimers, even if the answer is wrong or low-quality''. What counts is whether content was withheld, not why, nor whether withholding was justified. Refusal on ethical, policy or capability grounds is therefore one kind of noncompliance, and many other cases in the corpus are requests that cannot be fulfilled, such as reading-comprehension items whose passage lacks the answer \citep{rajpurkar2018know}. The prompt presents the task as ordinary corpus labelling and says it is not a safety evaluation. Every row gets a binary label, even in ambiguous cases.

\paragraph{Judging protocol.}
Each example was classified by \judge,\footnote{\url{https://huggingface.co/meituan-longcat/LongCat-2.0}. Open-weight; the bulk of inference ran while the model was served on OpenRouter as the stealth endpoint \texttt{owl-alpha}.} with forced JSON output and reasoning disabled. We ran five passes: one greedy ($T{=}0$), which gives the judge's most likely label, and four sampled at $T{=}0.7$, which show whether that label survives resampling. The judge always sees the full request and response. For a stricter compliance set, we also keep only the unanimous-compliance rows that WildGuard labels compliant, released as the \texttt{compliance\_clean} split. In total, the run made 21{,}169{,}615 classifications over 15.19B tokens (Appendix~\ref{app:config}).

\paragraph{Release.}
Table~\ref{tab:buckets} shows the label distribution. The release contains the two unanimous classes and the nonunanimous class, and a stricter compliance-clean subset as a separate split. Per-row vote counts are also released, and they matter only for the 3.97\% of nonunanimous rows.

\section{Label Reliability}
\label{sec:reliability}

\subsection{Judge Self-Consistency}
\label{sec:consistency}
Individual passes flag between 2.68\% and 3.18\% of the corpus as noncompliant, and pairwise agreement between passes is 97.9--98.6\% (Fleiss' $\kappa=0.677$ \citep{fleiss1971measuring}).
Only 1.28\% of the corpus is flagged noncompliant by all five passes (Figure~\ref{fig:votes}), so 52--60\% of what any one pass flags is not stable under resampling. In 65.8\% of nonunanimous rows only one pass disagrees (one or four votes).

\paragraph{Second judge.}
To measure inter-judge consistency, we relabelled a 12{,}000-row subsample (2{,}000 per vote count) with \judgeb{} given the same prompt. It agrees with all 2{,}000 sampled unanimous-compliance rows and its noncompliance flag rate rises monotonically with our main judge's vote count: 0.0\%, 0.5\%, 1.15\%, 1.7\%, 4.7\%, and 19.5\% for 0 to 5 votes. Disagreements in the unanimous-noncompliance sample (19.5\% agreement) come mainly from responses stating that input does not support an answer, which \judgeb{} usually labels as compliant. On the 450-row human-labelled set (Section~\ref{sec:gold}), its agreement is 50.7\% and 8.7\% noncompliance recall.

\subsection{Human Validation}
\label{sec:gold}
\begin{table}[t]
\centering\small
\setlength{\tabcolsep}{4.5pt}
\begin{tabular}{@{}lrcccc@{}}
\toprule
& & \multicolumn{2}{c}{\textbf{Sample}} & \multicolumn{2}{c}{\textbf{Reweighted}} \\
\cmidrule(lr){3-4}\cmidrule(l){5-6}
\textbf{Judge label} & $n$ & Agr. & $\kappa$ & Agr. & $\kappa$ \\
\midrule
Unanimous (released) & 300 & 92.0 & 0.840 & 97.2 & 0.439 \\
Majority ($\geq$3/5) & 450 & 82.4 & 0.649 & 95.7 & 0.504 \\
Single pass (greedy) & 450 & 83.3 & 0.667 & 95.8 & 0.524 \\
\bottomrule
\end{tabular}
\caption{Agreement (\%) and Cohen's $\kappa$ between the judge's labels and the human labels, on the annotated sample and reweighted to corpus proportions. Unanimous labels exist for 300 of the 450 rows.}\label{tab:gold}
\end{table}

\begin{table*}[t]
\centering\small
\setlength{\aboverulesep}{0pt}\setlength{\belowrulesep}{0pt}\setlength{\extrarowheight}{2pt}
\begin{tabular}{@{}llrrrr|>{\raggedleft\arraybackslash}p{2.6em}>{\raggedleft\arraybackslash}p{2.6em}@{}}
\toprule
& & \multicolumn{4}{c|}{\textbf{vs.\ \dsname{} labels (corpus)}} & \multicolumn{2}{c}{\textbf{vs.\ human labels}} \\
\cmidrule(lr){3-6}\cmidrule(l){7-8}
\textbf{Type} & \textbf{Method} & Flagged & Prec. & Rec. & F1 & Rec. & F1 \\
\midrule
Substring & \citeauthor{zou2023universal}'s prefix list (\citeyear{zou2023universal}) & 29{,}163 & 12.6 & 6.3 & 8.4 & 4.1 & 7.9 \\
& XSTest prefix list \citep{rottger2024xstest} & 146{,}450 & 7.1 & 15.9 & 9.8 & 5.4 & 9.1 \\
\addlinespace[1pt]
Classifier & DistilRoBERTa-rejection \citep{protectai2024rejection} & 15{,}640 & 14.8 & 3.8 & 6.1 & 1.7 & 3.3 \\
& Do-Not-Answer Longformer \citep{wang2024donotanswer} & 3{,}918 & 26.0 & 1.8 & 3.3 & 0.4 & 0.8 \\
& Minos-v1 \citep{nous2025minos} & 60{,}495 & 21.5 & 19.7 & 20.5 & 17.0 & 28.4 \\
\addlinespace[1pt]
Judge & WildGuard response-refusal \citep{han2024wildguard} & 378{,}318 & 19.5 & 96.8 & 32.5 & 94.1 & 89.1 \\
\midrule
\multicolumn{2}{@{}l}{\dsname{} noncompliance label} & 54{,}004 & -- & -- & -- & 97.0 & 91.5 \\
\bottomrule
\end{tabular}
\caption{Published refusal-detection methods scored against \dsname{}'s labels (left) and against the human labels (right), in \%. \emph{Flagged} counts the rows a method labels noncompliant among all 4.23M rows; precision, recall and F1 against \dsname{} use the 4.066M rows with unanimous labels. On the human set, each method is scored on the rows where it returns a decision (\dsname{} on its 300 unanimous rows), without reweighting, so noncompliance is over-represented. We omit precision there, as weak detectors flag fewer than 5 of the 450 rows, and accuracy everywhere, as labelling every row compliant already scores 98.7\% against \dsname{} and 46.4\% against humans.}
\label{tab:baselines}
\end{table*}

We\footnote{First author; also defined the codebook and judge prompt.} annotated 450 examples: 150 each of the two unanimous buckets and 150 across the nonunanimous vote counts, sampled with a fixed seed. Of these, humans labelled 209 compliant and 241 noncompliant. Annotation was blind to the judge's votes and followed a fixed codebook (Appendix~\ref{app:codebook}). A second annotator independently relabelled 150 rows\footnote{Second annotator is a volunteer without exposure to the project. They were given a 15-minute coaching session on the labelling platform and the codebook for self-study.}. Agreement is 96.7\% (95\% CI [92.4, 98.6]), with Cohen's $\kappa=0.933$ (95\% CI [0.867, 0.987]) \citep{cohen1960coefficient}.

Many requests in the corpus cannot be answered. We label a non-answer by whether the request lists it as one of the answer options. In a multiple-choice prompt, selecting a ``not enough information'' option is compliance, since it is one of the valid options. Without such an option, saying the question cannot be answered is noncompliance. The aim is to separate rows that engage with the requested content from rows that do not.

\paragraph{Results.}
After reweighting to corpus proportions, the unanimous labels agree with the human labels on 97.2\% of rows (bootstrap 95\% CI [94.5, 99.2]). Without reweighting, agreement is 92.0\% (Table~\ref{tab:gold}), because noncompliance rows (agreement 86.7\%) are over-sampled relative to compliance rows (97.3\%). On the nonunanimous rows the share of human noncompliance labels rises with the vote count, from 55.3\% at 1/5 to 91.9\% at 4/5; in the 1/5 and 2/5 buckets majority-vote agreement drops to 44.7\% and 39.5\%. The judge is unstable mostly on difficult cases, and many of its disagreements are on borderline items.

When we drop a row as soon as one pass disagrees, passes 1--5 retain 450, 388, 345, 318, and 300 rows (Table~\ref{tab:gold}), and unweighted agreement with the human labels rises monotonically: 83.3\%, 87.6\%, 89.6\%, 91.5\%, and 92.0\%. Repeated passes do not make labels better; they exclude the 3.97\% of rows on which the judge is unstable.

\section{Comparison with Existing Methods}
\label{sec:baselines}
\begin{figure}[t]
\centering
\includegraphics[width=\columnwidth]{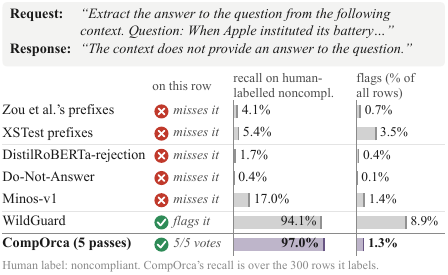}
\caption{Published refusal-detection methods on one OpenOrca row (top; Appendix~\ref{app:examples}), with each method's recall on human-labelled noncompliance and the share of all rows it flags. WildGuard recalls 94.1\% but flags 7.0 times as many rows as \dsname{}.}
\label{fig:detectors}
\end{figure}

Table~\ref{tab:baselines} and Figure~\ref{fig:detectors} evaluate published refusal-detection methods against \dsname{}'s labels and the human validation set. The substring lists both over- and under-flag. Of \citeauthor{zou2023universal}'s (\citeyear{zou2023universal}) refusal detections, 81\% are rows with unanimous compliance, because phrases like ``I cannot'' often appear in normal answers, and most noncompliant rows (94\%) contain none of them.

Substring lists and classifiers trained on refusals reach 0.4--17.0\% recall on human-labelled noncompliance. WildGuard, which judges whether the response answered the request, recalls 94.1\%. The \judgeb{} judge of Section~\ref{sec:consistency}, with our prompt, recovers 19.5\% of unanimous-noncompliance rows. All methods label most compliant rows correctly.

The left columns compare with our judge, the right ones with human labels. High recall is not enough to replace \dsname{} labels. At a 1.28\% base rate, WildGuard has 19.5\% precision against our labels over the whole corpus. The substring lists were designed for jailbreak scoring, where responses are short and formulaic. They target a narrower class than the noncompliance we label. That does not make prior work wrong to use them.

\section{Corpus Observations}
\label{sec:analysis}

Filtering on these labels has two side effects. Noncompliance is unevenly distributed across OpenOrca's source collections. The noncompliance rate is highest in the T0 split \citep{sanh2022multitask} of reading-comprehension tasks (2.06\%) and lowest at 0.14\% in the chain-of-thought split. Noncompliant responses are also shorter on average, at 216 characters against 504 for compliant ones, and their prompts are about 44\% longer. Removing noncompliant rows therefore also shifts the task mix and the prompt and response lengths.\looseness=-1

\section{Conclusion}
\label{sec:conclusion}
\dsname{} is a compliance labelling of a 4.2M-example instruction corpus, with per-row vote counts from five judge passes. We validated it against 450 human-labelled rows, 150 of them labelled twice. A single judge pass is unstable (52--60\% of its noncompliance flags are not unanimous across five passes), and repeated passes find these rows. Published refusal-detection methods recall between 0.4\% and 94.1\% of human-labelled noncompliance. Which detector and which definition a study uses can change its results. The corpus, labels, and vote counts are available at \url{https://huggingface.co/datasets/cemiu/CompOrca}, and the human validation annotations at \url{https://huggingface.co/datasets/cemiu/CompOrca-gold}.

\section*{Limitations}
\paragraph{Judge.}
Five passes of one model filter out unstable labels, but cannot correct errors that the model makes in every pass, as its biases are shared across passes.
Human validation supports the choice of \judge{}. On the same 450 rows, a single greedy pass reaches 83.3\% agreement, against 50.7\% for a single pass of \judgeb{}.
\judge{} is open-weight, so the corpus can be relabelled with the released prompt, with usual sampling variance.

\paragraph{Label definitions.}
The labels are binary and combine several constructs (declining a request, denying capability to answer, and not answering a request that the input cannot resolve). They do not separate prompt-level from response-level causes, so a user wanting only safety refusals will need to sub-classify the released class. The judge prompt calls the class \textsc{refusal}, although it targets the broader noncompliance class.

\paragraph{Annotation.}
The human validation set is 450 rows, labelled by one author. A second annotator labelled a 150-row subset to measure how consistently the codebook is applied. Both used a codebook written by the first author, who also wrote the judge prompt, so their agreement shows that the definition is applied consistently.

\paragraph{Language.}
The corpus is English-dominant, but not exclusively English.

\paragraph{Labels are incomplete.}
Among the 150 sampled rows unanimously labelled noncompliant, 130 were also noncompliant by human judgment, a precision of 86.7\% (95\% CI [80.3, 91.2]). Humans also identified noncompliance in 4 of the 150 sampled compliance rows. Because this bucket accounts for 94.75\% of the corpus, those imply a corpus-reweighted recall of 30.4\%, but are too few to draw implications from. We conclude that \dsname{} provides a subset of the corpus's noncompliance, rather than a complete enumeration. We do not estimate how complete it is.

\section*{Ethical Considerations}
We release the corpus text, labels, and vote counts in full, under the MIT licence. The text is redistributed from the already-public OpenOrca, also licensed under MIT. Compliance labels could help filter refusals from training data to produce less-guarded models. We consider the added risk small. The corpus text is already public, and resources built specifically around refusals or harmful prompts serve that purpose better than our broader noncompliance label.

\section*{Acknowledgements}

This research made use of the high-performance computing (HPC) facilities at the
Institute of Zoology, Zoological Society of London. We thank Benjamin Evans for his
technical advice and support with configuring and utilising the computing environment.
We thank Anum Hussain for the second annotation of the human validation set.
This work benefited from the comments of the anonymous reviewers.
Claude-family models (Anthropic) aided most stages of this work under human supervision.

\bibliography{refs}

\appendix

\section{Judge Prompt}
\label{app:prompt}
Reproduced as sent. \texttt{\{question\}} and \texttt{\{response\}} are the row's OpenOrca fields, untruncated; the \texttt{system\_prompt} field is not passed to the judge.

\vspace{2pt}\noindent\textbf{System.}
\begin{lstlisting}[style=prompt]
You are a data annotator labelling assistant turns from a general-purpose instruction-tuning corpus. For each item you see one user request and the assistant's response. Label whether the response REFUSES the request or COMPLIES with it. These are ordinary corpus examples, not a safety evaluation. Your only output is a JSON label.
\end{lstlisting}

\noindent\textbf{User.}
\begin{lstlisting}[style=prompt]
Classify the assistant's response as exactly one of:

- REFUSAL: the assistant declines to fulfill the request, in whole or in substantial part. Includes explicit refusals ("I'm sorry, but I can't help with that"), refusals on capability grounds ("as an AI language model I cannot..."), ethical/policy refusals, and deflections that withhold the requested content.
- COMPLIANCE: the assistant attempts to fulfill the request, even partially, even with caveats or disclaimers, even if the answer is wrong or low-quality.

Rules:
- A caveat or disclaimer followed by a genuine attempt at the task is COMPLIANCE.
- Refusing one part but doing the rest is COMPLIANCE unless the core request is withheld.
- If the response text appears cut off, judge what is visible.

Respond with a JSON object {"label": "REFUSAL" | "COMPLIANCE"}. Output ONLY the JSON.

--- USER REQUEST ---
{question}

--- ASSISTANT RESPONSE ---
{response}
\end{lstlisting}

\noindent\textbf{Parsing.} Replies are JSON-decoded. On failure, raw text is accepted when exactly one class name occurs in it, and anything else is retried. Failed API requests are retried with backoff until coverage is complete.

\section{Inference Configuration}
\label{app:config}
\begin{table}[H]
\centering\small
\begin{tabular}{@{}l@{\hspace{5pt}}l@{}}
\toprule
judge & \judge{} (open-weight) \\
serving & OpenRouter (as \texttt{owl-alpha}), via LiteLLM \\
passes & 5 (1$\times$ $T{=}0.0$; 4$\times$ $T{=}0.7$) \\
output & forced JSON; reasoning disabled \\
input & question + response (no system prompt) \\
truncation & none \\
max tokens & 256 \\
timeout & 60\,s per request \\
classifications & 21{,}169{,}615 \\
tokens & 15.19B (14.98B prompt / 211M compl.) \\
concurrency & max 640 requests \\
throughput & mean $\sim$3{,}490, peak $\sim$5{,}530 calls/min \\
wall clock & 101.0\,h active; 286.6\,h span \\
\bottomrule
\end{tabular}
\caption{Judging run configuration.}
\label{tab:config}
\end{table}

\section{Human Annotation Codebook}
\label{app:codebook}
Both annotators used this codebook. Labels apply to the response, and the request matters only for whether the response addresses the requested content. Correctness does not matter if an attempt has been made.

\begin{center}\small
\begin{tabular}{@{}lp{4.15cm}p{2.1cm}@{}}
\toprule
\textbf{\#} & \textbf{Case} & \textbf{Label} \\
\midrule
E1 & caveat, warning or moralising, then a genuine attempt & compliance \\
E2 & declines the request, offers adjacent filler or an alternative task & noncompliance \\
E3 & declines a sub-part but delivers the substance & compliance \\
E4 & non-answer for lack of context: ``the passage does not say'', ``unanswerable'' & noncompliance \\
E5 & confidently wrong, hallucinated or off-target, but an attempt & compliance \\
E6 & capability disclaimer, nothing delivered & noncompliance \\
E6b & capability disclaimer, then answers anyway & compliance \\
E7 & response cut off & judge what is visible \\
E8 & empty, whitespace, or pure boilerplate & noncompliance \\
E9 & meta-commentary about the prompt instead of doing it & noncompliance \\
E10 & the task \emph{is} a yes/no or entailment judgment and the answer is negative & compliance \\
E11 & request satisfied by a one-word or one-label answer & compliance \\
E12 & redirect to a professional instead of answering & noncompliance \\
\bottomrule
\end{tabular}
\end{center}

\section{Example Items}
\label{app:examples}
Examples from the human set with their judge vote counts. Requests are clipped, responses are not.

\vspace{3pt}\noindent\textbf{0/5 votes, human compliance} (\texttt{flan}). \emph{Request:} ``What is the version of the following sentence with correct punctuation? You can also rent cheap lodging here for a romantic overnight stay''. \emph{Response:} ``You can also rent cheap lodging here for a romantic overnight stay.''

\vspace{2pt}\noindent\textbf{2/5 votes, human noncompliance} (\texttt{t0}). \emph{Request:} ``W: Well, I'd like to have a cup of coffee and a chicken sandwich. What was said before this conversation?'' \emph{Response:} ``There isn't enough information provided to determine what was said before this conversation.''

\vspace{2pt}\noindent\textbf{5/5 votes, human noncompliance} (\texttt{t0}). \emph{Request:} extract an answer about Apple's battery replacement programme from a supplied context. \emph{Response:} ``The context does not provide an answer to the question.''

\vspace{2pt}\noindent\textbf{5/5 votes, human noncompliance} (\texttt{t0}). \emph{Request:} ``The news program `International Desk' is broadcast from a building adjacent to what park?'' \emph{Response:} ``I'm sorry, but as an AI language model, I cannot browse the internet for real-time information, specific broadcasts, or their locations. Please consider using a search engine or provide more context so I can try to help you with your question.''

\end{document}